\documentclass{article}
\PassOptionsToPackage{numbers, compress}{natbib}
\usepackage[preprint]{neurips_2026}

\usepackage[utf8]{inputenc}
\usepackage[T1]{fontenc}
\usepackage{hyperref}
\usepackage{url}
\usepackage{microtype}
\usepackage{graphicx}
\usepackage{subcaption}
\usepackage{booktabs}

\usepackage{amsmath}
\usepackage{amssymb}
\usepackage{amsfonts}
\usepackage{mathtools}
\usepackage{amsthm}
\usepackage{multirow}
\usepackage[table]{xcolor}
\usepackage{xspace}
\usepackage{makecell}
\usepackage{bm}
\usepackage{nicefrac}
\usepackage{algorithm}
\usepackage{algorithmic}
\usepackage{enumitem}
\usepackage{listings}

\newcommand{\ourmethod}{\textsc{AEvo}\xspace}
\newcommand{\method}{\ourmethod}

\newcommand{\yuyu}[1]{}
\newcommand{\didi}[1]{}
\newcommand{\bang}[1]{}

\definecolor{rowgray}{gray}{0.85}
\definecolor{rowlightblue}{RGB}{215, 230, 255}
\definecolor{codebg}{RGB}{247, 247, 247}

\definecolor{geminiBG}{RGB}{235, 245, 255}
\definecolor{geminiDark}{RGB}{0, 80, 180}

\definecolor{deepseekBG}{RGB}{235, 250, 235}
\definecolor{deepseekDark}{RGB}{0, 100, 0}

\lstdefinestyle{papercode}{
  basicstyle=\ttfamily\scriptsize,
  backgroundcolor=\color{codebg},
  breaklines=true,
  breakatwhitespace=false,
  columns=fullflexible,
  frame=single,
  framerule=0pt,
  keepspaces=true,
  showstringspaces=false,
  xleftmargin=0.6em,
  xrightmargin=0.2em,
  framexleftmargin=0.6em,
  aboveskip=0.6em,
  belowskip=0.6em
}

\usepackage[capitalize,noabbrev]{cleveref}

\theoremstyle{plain}

\theoremstyle{definition}

\theoremstyle{remark}

\usepackage[disable,textsize=tiny]{todonotes}

\title{Harnessing Agentic Evolution}

\author{%
    \textbf{Jiayi Zhang}$^{1,2}$,
    \textbf{Yongfeng Gu}$^{2}$,
    \textbf{Jianhao Ruan}$^{1,2}$,
    \textbf{Maojia Song}$^{3}$,
    \textbf{Yiran Peng}$^{2}$, \\
    \textbf{Zhiguang Han}$^{4}$,
    \textbf{Jinyu Xiang}$^{1}$,
    \textbf{Zhitao Wang}$^{5}$,
    \textbf{Caiyin Yang}$^{6}$, 
    \textbf{Yixi Ouyang}$^{2}$,\\ 
    \textbf{Bang Liu}$^{7}$,
    \textbf{Chenglin Wu}$^{2,\dagger}$,
    \textbf{Yuyu Luo}$^{1,\dagger}$
    \\
    $^1$The Hong Kong University of Science and Technology (Guangzhou),
    $^2$DeepWisdom, \\
    $^3$Singapore University of Technology and Design,
    $^4$Nanyang Technological University, \\
    $^5$Shanghai Jiao Tong University,
    $^6$Tsinghua University,
    $^7$Université de Montréal \& Mila
  }

\begin{document}

\maketitle

\begin{abstract}

Agentic evolution has emerged as a powerful paradigm for improving programs, workflows, and scientific solutions by iteratively generating candidates, evaluating them, and using feedback to guide future search.
However, existing methods are typically instantiated either as fixed hand-designed procedures that are modular but rigid, or as general-purpose agents that flexibly integrate feedback but can drift in long-horizon evolution.
Both forms accumulate rich evidence over time, including candidates, feedback, traces, and failures, yet lack a stable interface for organizing this evidence and revising the mechanism that drives future evolution.
We address this limitation by formulating \textbf{agentic evolution as an interactive environment}, where the accumulated evolution context serves as a process-level state.
We introduce \method, a harnessed meta-editing framework in which a meta-agent observes this state and acts not by directly proposing the next candidate, but by editing the procedure or agent context that controls future evolution.
This unified interface enables \method to steer both procedure-based and agent-based evolution, making accumulated evidence actionable for long-horizon search.
Empirical evaluations on agentic and reasoning benchmarks show that \method outperforms five evolution baselines, achieving a \textbf{26\%} relative improvement over the strongest baseline.
Across three open-ended optimization tasks, \method further outperforms four evolution baselines and achieves state-of-the-art performance under the same iteration budget.
\end{abstract}
\section{Introduction}
\label{sec:introduction}

Agentic evolution reframes LLM-based problem solving as a process of constructing and revising solutions~\citep{liu2025advances, gao2025evolvesurvey}. 
Instead of treating the model only as a generator of candidate answers, these methods use LLMs, agentic workflows, or coding agents to drive iterative improvement: produce candidate artifacts, interpret feedback from evaluation, and influence what the system explores next~\citep{qu2026coral}. 
This paradigm has been applied to program synthesis~\citep{jimenez2023swe}, scientific discovery~\citep{lu2024ai, yuksekgonul2026learning}, systems optimization~\citep{tbench_2025, deng2025interactcomp}, and agent self-improvement~\citep{zhang2026hyperagents, wang2025huxley, ruan2026aorchestra}. 
In this paper, we use \textbf{agentic evolution} to broadly refer to evolution processes whose search behavior is driven by either structured agentic procedures or general-purpose agents.

Existing agentic evolution methods typically instantiate this paradigm in two ways.
In \textbf{procedure-based evolution}, a predefined outer loop controls parent selection, candidate generation, evaluation, and population update~\citep{novikov2025alphaevolve, ye2026evaluation, zhang2024aflow}.
This makes evolution modular and reproducible, but also ties long-horizon search to fixed selection rules, feedback summaries, and update heuristics.
In \textbf{agent-based evolution}, a general-purpose agent manages the search process by observing feedback, inspecting traces, editing candidates, writing tools, and deciding what to try next~\citep{karpathy2026autoresearch, qu2026coral}.
This gives evolution greater flexibility, but the agent can drift as candidates, logs, hypotheses, and intermediate files accumulate.
In both cases, long-horizon evolution remains prone to local optima: procedures may repeatedly exploit the same hand-designed search pattern, while agents may overcommit to misleading evidence or stale assumptions in a growing context.

Recent work has tried to address these limitations by either broadening exploration with collaborative agents~\citep{qu2026coral} or making the evolution mechanism self-modifying~\citep{zhang2026hyperagents}.
These directions show that stronger search context and editable improvement mechanisms are useful, but they do not by themselves provide a stable interface for long-horizon evolution.
The core challenge is that evolution accumulates candidates, feedback, traces, failures, and intermediate decisions over time, yet lacks a unified way to organize this evidence and revise the mechanism that drives future evolution.

\begin{figure}[!t]
\centering
\includegraphics[width=\linewidth]{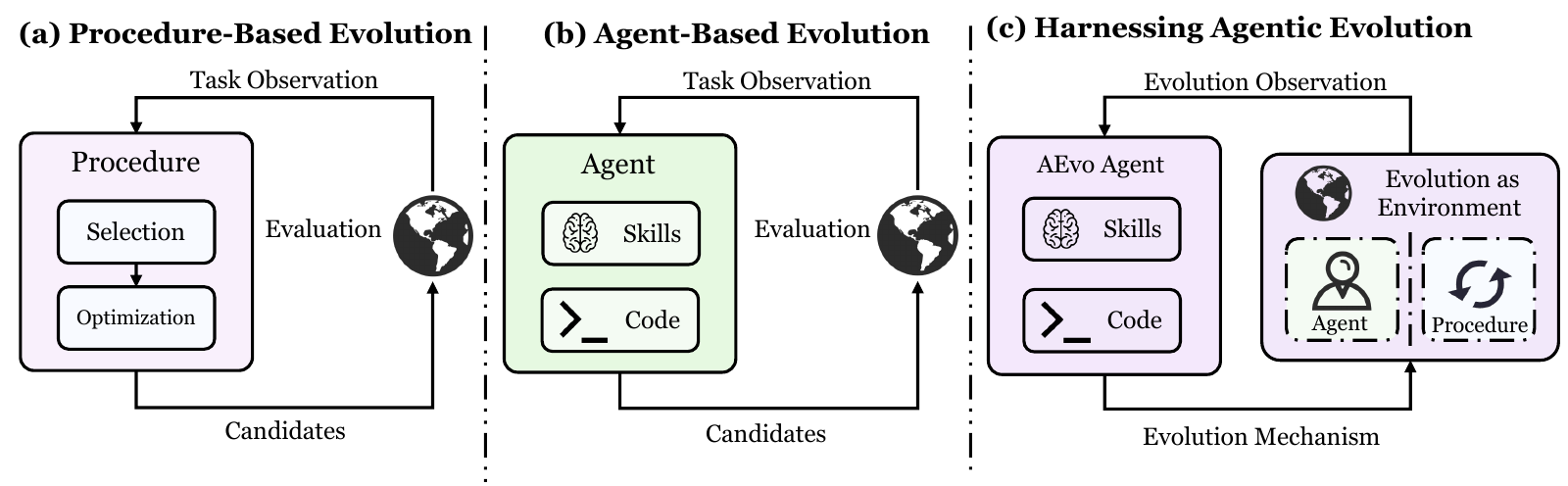}
\caption{
\textbf{Harnessing agentic evolution as an interactive environment.}
(a) Procedure-based evolution runs a fixed loop for selection, optimization, evaluation, and update.
(b) Agent-based evolution lets a general-purpose agent manage search through feedback, tools, skills, and code actions.
(c) \method treats the evolution process as an interactive environment.
The accumulated evolution context becomes process-level state, while a meta-agent edits the underlying procedure or agent operating context that controls future evolution.
}
\label{fig:teaser}
\end{figure}

We address this challenge by formulating \textbf{agentic evolution as an interactive environment}.
As illustrated in Figure~\ref{fig:teaser}, this view shifts evolution from an unstructured iterative process into an environment that exposes process-level state and supports external intervention.
The state is the accumulated evolution context, including candidates, feedback, traces, failures, costs, and search history.
The transition mechanism is the current evolution mechanism: either an explicit search procedure or the operating context that shapes a general-purpose agent's future decisions.
A meta-agent acts on this environment not by generating the next candidate, but by editing the mechanism that controls how future evolution proceeds.
This makes the same environment view applicable to both hand-designed procedures and general-purpose evolution agents.

Realizing this view requires a harnessed design.
The evolution environment is large, noisy, and constantly changing.
Without a stable interface, a meta-agent may lose track of reliable evidence, revisit old attempts or make edits whose effects are hard to verify.
At the same time, evaluation and candidate records must remain protected from the agents that modify the evolution process.
These challenges motivate a harness that makes evolution observable, editable, and externally governed.

We therefore introduce \method, a harnessed framework for meta-editing agentic evolution.
\method standardizes the evolution workspace, protects the evaluator, records every evaluated candidate into a searchable history, and exposes process-level information to the meta-agent.
It then runs evolution through a two-phase loop. 
In the meta-editing phase, the meta-agent edits the current mechanism and specifies how the next segment should run.
In the evolution segment, the updated mechanism runs under this plan and produces multiple candidates before the next meta-agent intervention.
The same loop can revise both procedure- and agent-based evolution, reducing the risk of local optima.

Our contributions are threefold.
(1) \textbf{Environment Formulation:} We formulate agentic evolution as an interactive environment, where accumulated evolution context becomes process-level state and meta-actions edit the mechanism that drives future evolution.
(2) \textbf{Harnessed Meta-Editing:} We introduce \method, a harnessed framework for meta-editing agentic evolution that protects evaluation, records evaluated candidates, and supports coarse-grained intervention through meta-editing phases and evolution segments.
(3) \textbf{Cross-Form Instantiation and Evaluation:} We instantiate the same framework on both procedure-based and agent-based evolution, showing that \method can revise either explicit procedure components or agent operating contexts.
On standard agentic and reasoning benchmarks, \method outperforms five evolution baselines and achieves a 26\% relative improvement over the strongest baseline.
On three open-ended optimization tasks, \method outperforms four evolution baselines and achieves state-of-the-art performance under the same iteration budget.
\section{Related Work}
\label{sec:related_work}

\textbf{Agentic Evolution.}
A growing line of work uses LLMs and agents to iteratively improve artifacts through generation, feedback, and revision~\citep{DBLP:conf/icml/Li0FXC0L25,li2026deepeye,DBLP:journals/corr/abs-2510-17586,wu2026autowebworldsynthesizinginfiniteverifiable}. 
Prompt methods optimize language-model programs or feedback-driven prompts, including DSPy~\citep{khattab2024dspy}, SPO~\citep{xiang2025self}, TextGrad~\citep{yuksekgonul2024textgrad}, and GEPA~\citep{agrawal2025gepa}. 
Another line studies the automated design and evolution of agentic systems and workflows, such as ADAS~\citep{hu2024automated}, Darwin G\"odel Machine~\citep{zhang2025darwin}, Huxley-G\"odel Machine~\citep{wang2025huxley}, AFlow~\citep{zhang2024aflow}, RobustFlow~\citep{xu2025robustflow}, and SkillRL~\citep{xia2026skillrl}. 
Recent open-ended discovery systems further apply evolutionary search to scientific and algorithmic discovery, including AlphaEvolve~\citep{novikov2025alphaevolve}, OpenEvolve~\citep{openevolve}, TTS-Discover~\citep{yuksekgonul2026learning}, CORAL~\citep{qu2026coral}, SimpleTES~\citep{ye2026evaluation}, and ASI-Evolve~\citep{xu2026asi}. 
However, their search behavior is typically controlled either by fixed procedures or by agents directly managing candidate generation. 
In contrast, \method treats the evolution process itself as an interactive environment and studies how to steer the mechanism that controls future search.

\textbf{Agentic Meta-Evolution.}
Early meta-learning work showed that the learning rule itself can be optimized, for example by learning recurrent reinforcement-learning dynamics~\citep{wang2016learning} or evolving policy-gradient objectives~\citep{houthooft2018evolved}.
Recent agentic systems extend this idea to editable agent programs and memory systems.
HyperAgents study self-referential agent programs in which both task-solving behavior and the meta-improvement mechanism can be modified~\citep{zhang2026hyperagents}.
MemEvolve and ALMA similarly explore meta-evolution over agent memory designs~\citep{zhang2025memevolve, xiong2026learning}.
Unlike HyperAgents, which internalize meta-improvement within a self-modifying agent program, \method treats agentic evolution as an interactive environment observed and edited through an external harness, covering both hand-designed procedures and general-purpose agents while keeping evaluation and candidate recording externally governed.

\section{Problem Formulation}
\label{sec:formulation}

\subsection{Agentic Evolution}
\label{sec:agentic_evolution}

We formulate agentic evolution as a process for optimizing an artifact through repeated improvement rounds. 
Let $x \in \mathcal{X}$ denote the object being optimized, such as a program, prompt, workflow, skill, tool, or agent component. 
We use $r$ to index the evolution round. 
Each round produces a round context $c_r$, which contains the candidates generated in that round, their evaluation results, execution traces, failures, costs, and any intermediate information produced during optimization. 
The accumulated evolution context after $r$ rounds is denoted as
\[
\mathcal{C}_r = (c_1, c_2, \ldots, c_r).
\]
Finally, let $\Pi$ denote the optimization mechanism that advances evolution:
\[
c_r = \Pi(\mathcal{C}_{r-1}), 
\qquad 
\mathcal{C}_r = \mathcal{C}_{r-1} \oplus c_r,
\]
where $\oplus$ appends the newly produced round context to the accumulated evolution context. 
$\Pi$ does not have to be a fixed algorithm; it can also be an agentic process that reads the history, reasons over feedback, and decides how to generate the next candidate. 
Thus, $\Pi$ represents the mechanism by which search is continued from the current evolution context.

Under this formulation, existing agentic evolution methods mainly differ in how $\Pi$ is instantiated. 
In \textbf{procedure-based evolution }, $\Pi$ is a predefined outer loop whose behavior is mainly determined by selection and optimization: the selection rule chooses previous candidates or contexts from $\mathcal{C}_{r-1}$, while the optimization operator generates new candidates from the selected information. 
Evaluation assigns scores, traces, and feedback to the generated candidates, providing signals for future selection and update. 
In \textbf{agent-based evolution}, $\Pi$ is instead implemented by a general-purpose agent. 
Rather than following fixed selection-and-optimization rules, the agent reads the accumulated context $\mathcal{C}_{r-1}$ and decides what to do next, such as inspecting feedback, comparing candidates, modifying artifacts, writing tools, or generating new attempts. 
Thus, procedure-based evolution  specifies search control explicitly but rigidly, while agent-based evolution leaves search control implicit in the agent's context-conditioned behavior.

In both cases, evolution proceeds by repeatedly applying $\Pi$ while accumulating context $\mathcal{C}_r$. 
This context records not only the candidates produced by evolution, but also how search has unfolded through evaluation results, traces, failures, costs, and intermediate artifacts. 
The next subsection uses this accumulated context to define an environment view of evolution.

\subsection{Evolution as an Interactive Environment}
\label{sec:evolution_as_environment}

We treat the evolution process itself as an interactive environment for a meta-agent. 
At round $r$, the state of this environment is defined by the round index and the accumulated evolution context:
\[
s_r = (r, \mathcal{C}_r).
\]
When the optimization mechanism may change across rounds, we write the current mechanism as $\Pi_r$. 
This mechanism specifies the transition of the environment. 
Without intervention, the next round is produced by applying the current mechanism to the current context:
\[
c_{r+1} = \Pi_r(\mathcal{C}_r),
\qquad
s_{r+1} = (r+1, \mathcal{C}_r \oplus c_{r+1}).
\]
Thus, $\Pi_r$ is the transition rule that determines how the evolution process continues.

To interact with this environment, we introduce a meta-agent $M$. 
The role of $\Pi_r$ is to continue the candidate search, while the role of $M$ is to act on the evolution process that governs this search. 
Since the full state $s_r$ can be large and noisy, the meta-agent receives an observation extracted from the state:
\[
o_r = \Phi(s_r) = \Phi(r, \mathcal{C}_r),
\]
where $\Phi$ summarizes relevant information from the accumulated context, such as progress, repeated failures, invalid attempts, cost patterns, or redundant search directions.

Given this observation, the meta-agent produces an edit action:
\[
a_r = M(o_r).
\]
The action does not directly become the next candidate. 
Instead, it modifies the transition rule of the evolution environment:
\[
\Pi_{r+1} = \mathrm{Edit}(\Pi_r, a_r).
\]
The edited mechanism is then used to continue evolution:
\[
c_{r+1} = \Pi_{r+1}(\mathcal{C}_r),
\qquad
\mathcal{C}_{r+1} = \mathcal{C}_r \oplus c_{r+1}.
\]
In this sense, we formulate agentic evolution as an environment in which the state is the accumulated evolution context, the observation is a summary of this context, and the action edits the mechanism that controls future search.

This formulation applies to both forms of agentic evolution. 
For procedure-based evolution , editing $\Pi_r$ changes explicit components such as selection, optimization, feedback use, budget allocation, or update rules. 
For agent-based evolution, editing $\Pi_r$ changes the agentic context that shapes future decisions, such as skills, goals, tools, feedback format, or execution context. 
In both cases, the meta-agent steers evolution not by proposing one more candidate, but by modifying how subsequent search is carried out. 
Section~\ref{sec:methodology} describes the system design used to instantiate this formulation.

\section{Methodology}
\label{sec:methodology}

\begin{figure}[!t]
\centering
\includegraphics[width=\linewidth]{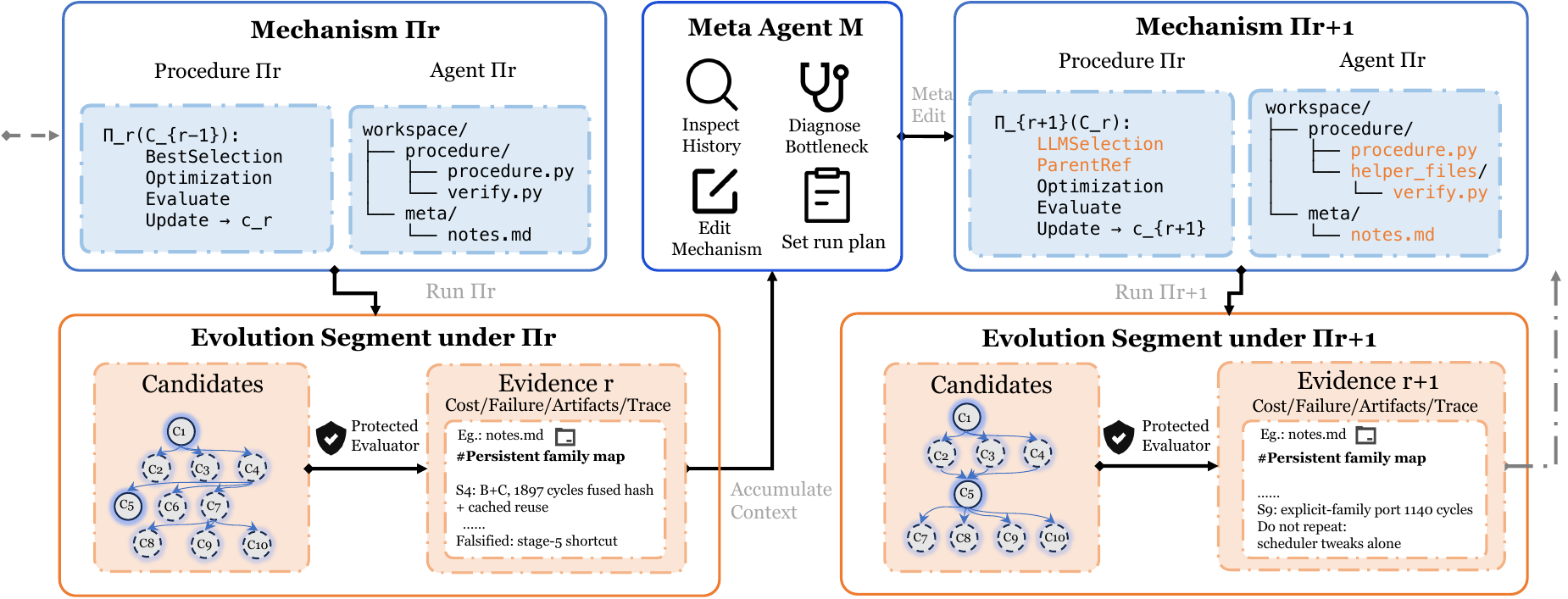}
\caption{
\textbf{Architecture of \method.}
The harness runs evolution segments under the current mechanism $\Pi_r$, protects evaluation, and records structured evidence.
A meta-agent observes this evidence to edit $\Pi_r$ into $\Pi_{r+1}$ and set the next run plan, enabling coarse-grained intervention over both procedures and agent contexts.
}
\label{fig:main_figure}
\end{figure}

Figure~\ref{fig:main_figure} illustrates \method.
\method instantiates the environment view in Section~\ref{sec:formulation} as a harnessed loop that alternates between a \emph{meta-editing phase} and an \emph{evolution segment}.
In the meta-editing phase, the meta-agent updates the current evolution mechanism $\Pi_r$ and specifies how the next segment should run, including its iteration budget and stopping conditions.
In the evolution segment, the updated mechanism runs under this plan and may produce multiple evaluated candidates before the next meta-agent intervention.
Thus, one meta-edit can govern a segment of future evolution rather than a single candidate.

\subsection{Design of \method}
\label{sec:design}

\textbf{Meta-editing phase.}
The meta-editing phase decides both \textbf{what to change} and \textbf{how to continue}.
The meta-agent can be any coding-capable agent that can inspect a workspace, edit files, execute commands, and follow the \method meta-agent skill specification, such as Claude Code~\citep{claudecode2025}, Codex~\citep{codex2025}, or open-source coding agents~\citep{yang2024sweagent, opencode2025}.
Given the current workspace, it inspects the accumulated history, then produces a meta-action consisting of a workspace edit and a run plan.
The workspace edit modifies files that define $\Pi_r$, such as procedure code, prompts, skills, goals, tools, feedback formats, validators, notes, or execution context.
The run plan specifies how the next evolution segment should proceed, including the allowed iterations, budget use, and stopping conditions.

This design makes the meta-agent a process-level editor rather than a candidate generator: it changes the mechanism and conditions under which future candidates are produced.
When evolution is productive, the meta-agent may allocate more iterations to the current mechanism; when it repeatedly produces invalid candidates, redundant attempts, or irrelevant exploration, the meta-agent may stop the segment and revise $\Pi_r$ before continuing.
The full meta-agent skill specification and pseudocode of the meta-editing loop are provided in Appendix~\ref{app:meta_agent_skill}.

\textbf{Harnessed evolution segment.}
An evolution segment is the interval executed after a meta-edit.
It runs the current mechanism $\Pi_r$ under the run plan produced by the meta-agent.
Depending on the setting, this segment may consist of several rounds of a procedure, or an inner-agent session that produces multiple candidate attempts.
Each candidate submitted for official evaluation passes through the harness-controlled evaluator, and the resulting artifact, score, trace, failure information, cost, and provenance are appended to the candidate history.

The harness provides the stable boundary needed for reliable meta-editing.
It organizes candidates, logs, traces, evaluation records, meta-agent instructions, and editable evolution components into a fixed workspace layout.
To prevent reward hacking, the evaluator is isolated from both the evolution agent and the meta-agent: agents can submit candidates, but they cannot inspect evaluator internals, access hidden benchmark artifacts, or directly write official scores.
The harness further exposes a command-line interface for initializing workspaces, launching evolution segments, inspecting recent status and candidate history, and continuing the current process.
Thus, the harness does not decide how evolution should improve; it provides the protected and inspectable interface through which evolution can be observed, edited by the meta-agent, recorded, and resumed.

\subsection{Instantiating \method}
\label{sec:instantiation}

\method applies the same two-phase loop to both forms of agentic evolution.
The outer loop is unchanged: the meta-agent edits the current mechanism $\Pi_r$ and specifies how the next evolution segment should run.
The difference lies in what $\Pi_r$ consists of.

\textbf{Procedure-based evolution.}
For procedure-based evolution, $\Pi_r$ is an explicit evolution procedure.
It defines how previous candidates or contexts are selected, how new candidates are generated, how evaluation feedback is used, and how candidate history is updated.
A meta-action therefore edits the procedure itself, such as revising the selection strategy, changing the optimization operator, altering the feedback summary, adding local filtering or retry logic, adjusting budget use, or repairing candidate management.
The edited procedure then controls the next evolution segment, which may run for multiple rounds before the next meta-agent intervention.

\textbf{Agent-based evolution.}
For agent-based evolution, $\Pi_r$ is the operating context of a general-purpose evolution agent, including goals, skills, tools, memory files, shared notes, validators, and execution setup.
A meta-action therefore edits the conditions under which the next inner-agent session will evolve, such as revising a skill, rewriting the session goal, changing how evaluator feedback is presented, or reorganizing shared notes.
The inner agent remains responsible for generating candidates, while \method revises the context that shapes future evolution.
\section{Experiments}

\subsection{Settings}

\textbf{Tasks.}
We evaluate \ourmethod on two standard benchmarks, Terminal-Bench~\citep{tbench_2025} and ARC-AGI-2~\citep{chollet2025arc}, and three open-ended optimization tasks, circle\_packing\_26 (CP26)\citep{cp26}, autocorrelation\_second (AC2)\citep{ac2}, and Anthropic's Kernel optimization task~\citep{takehome}. Together, these tasks cover agentic problem solving, abstract reasoning, and code-evolving open-ended optimization. Detailed task definitions and evaluation protocols are given in Appendix~\ref{appendix:task_metric_details}.

\textbf{Baselines.}
We group baselines into three families: single-agent inference, agent-based evolution, and procedure-based evolution. On Terminal-Bench and ARC-AGI-2, we compare against one-shot ReAct~\citep{yao2022react} and five procedure-based evolution baselines: ADAS~\citep{hu2024automated}, DGM~\citep{zhang2025darwin}, AFlow~\citep{zhang2024aflow}, SPO~\citep{xiang2025self}, and GEPA~\citep{agrawal2025gepa}. On the three open-ended tasks, we compare against two agent-based evolution baselines, Codex and Claude Code, and two procedure-based evolution baselines, OpenEvolve~\citep{openevolve} and HyperAgents~\citep{zhang2026hyperagents}. This setup lets us compare both variants of \ourmethod against systems that either keep a fixed search procedure or rely on an agent to improve artifacts directly.

\textbf{Implementation Details.}
\ourmethod is instantiated in two forms. In the procedure-based setting, a meta-agent edits the evolution procedure while leaving the task evaluator fixed. In the agent-based setting, the meta-agent steers a coding-agent harness through prompts, notes, and reusable utilities. We use Claude Code and Codex as the meta-agent interfaces, backed by Claude-Opus-4.7 and GPT-5.4 as the optimization models. For Terminal-Bench and ARC-AGI-2, candidate execution uses Gemini-3-Flash. All models are accessed through APIs. Full hyperparameter settings, round budgets, early-stopping criteria, and initialization details are given in Appendix~\ref{appendix:impl_details}.

\textbf{Metrics.}
For Terminal-Bench and ARC-AGI-2, we report task score and the first optimization round that reaches the best score. Results are summarized with Avg@3 over three independent runs. For the open-ended tasks, we report the task-native objective with Best@3 over three runs, together with the first round that reaches the best result (Best R.) and the average dollar cost per optimization round (\$/R). Exact task-specific objectives and cost computation are provided in Appendix~\ref{appendix:task_metric_details}.

\subsection{Main Results}

\begin{table*}[htbp]
\caption{
Open-ended optimization results.
Task 1/2/3 correspond to circle\_packing\_26, autocorrelation\_second, and Kernel optimization task.
Scores denote Best@3 over three runs using the model shown in the Model column, Best R. denotes the first round reaching the reported best score, and \$/R denotes the average cost per optimization round.
Arrows indicate optimization direction; \textbf{bold} and \underline{underline} mark the best and second-best scores.
}
\label{tab:open-ended-results}
\centering
\footnotesize
\setlength{\tabcolsep}{1.9pt}
\renewcommand{\arraystretch}{1.15}

\begin{tabular}{lll ccccccccc}
\toprule[1.2pt]
\multirow{2}{*}[-3pt]{\textbf{Category}}
& \multirow{2}{*}[-3pt]{\textbf{Method}}
& \multirow{2}{*}[-3pt]{\textbf{Model}}
& \multicolumn{3}{c}{\textbf{Task 1 $\uparrow$}}
& \multicolumn{3}{c}{\textbf{Task 2 $\uparrow$}}
& \multicolumn{3}{c}{\textbf{Task 3 $\downarrow$}} \\
\cmidrule(lr){4-6}
\cmidrule(lr){7-9}
\cmidrule(lr){10-12}
& &
& \textbf{Score} & \textbf{Best R.} & \textbf{\$/R}
& \textbf{Score} & \textbf{Best R.} & \textbf{\$/R}
& \textbf{Score} & \textbf{Best R.} & \textbf{\$/R} \\
\midrule[0.8pt]

\multirow{2}{*}{\makecell{Agent-Based\\Evolution}}
& Codex       & GPT-5.4
& \textbf{2.6359} & 3 & 0.82
& 0.9176 & 96 & 0.04
& 1667 & 4 & 0.96 \\
& Claude Code & Claude-Opus-4.7
& \underline{2.6305} & 50 & 0.78
& \underline{0.9438} & 44 & 0.81
& 1615 & 97 & 0.51 \\

\midrule[0.3pt]

\multirow{4}{*}{\makecell{Procedure-Based\\Evolution}}
& OpenEvolve  & Claude-Opus-4.7
& 2.6303 & 80 & 0.42
& 0.9186 & 99 & 0.67
& 2411 & 99 & 0.62 \\
& HyperAgents & Claude-Opus-4.7
& \textbf{2.6359} & 32 & 9.50
& 0.9245 & 48 & 2.83
& 7086 & 86 & 1.56 \\
& OpenEvolve  & GPT-5.4
& 2.6341 & 19 & 0.23
& 0.9118 & 74 & 0.54
& 2464 & 100 & 0.57 \\
& HyperAgents & GPT-5.4
& 2.6359 & 47 & 3.19 
& 0.9237 & 61 & 1.46
& 3015 & 98 & 1.03 \\

\midrule[0.3pt]

\multirow{3}{*}{Ours}
& \ourmethod$_{\texttt{Procedure}}$ & Claude-Opus-4.7
& \textbf{2.6359} & 4 & 1.47
& 0.9278 & 29 & 0.70
& 1803 & 55 & 1.37 \\
& \ourmethod$_{\texttt{Agent}}$ & Claude-Opus-4.7
& \textbf{2.6359} & 2 & 0.34
& \textbf{0.9459} & 99 & 1.40
& \underline{1519} & 55 & 1.27 \\
& \ourmethod$_{\texttt{Agent}}$ & GPT-5.4
& \textbf{2.6359} & 17 & 0.32
& 0.9398 & 100 & 1.31
& \textbf{1138} & 99 & 1.23 \\

\bottomrule[1.2pt]
\end{tabular}

\vspace{0.4em}
\end{table*}

\begin{table*}[htbp]
\caption{
Standard benchmark results.
Scores denote Avg@3 over three runs with Gemini-3-Flash as the execution model, and Best R. denotes the first round reaching the reported best score.
The \textbf{bold} and \underline{underline} mark the best and second-best scores.
}
\label{tab:standard-results}
\centering
\small
\setlength{\tabcolsep}{4pt}
\renewcommand{\arraystretch}{1.15}

\begin{tabular}{lll cccccc}
\toprule[1.2pt]
\multirow{2}{*}[-3pt]{\textbf{Category}}
& \multirow{2}{*}[-3pt]{\textbf{Method}}
& \multirow{2}{*}[-3pt]{\textbf{Model}}
& \multicolumn{2}{c}{\textbf{Terminal-Bench $\uparrow$}}
& \multicolumn{2}{c}{\textbf{ARC-AGI-2 $\uparrow$}} \\
\cmidrule(lr){4-5}
\cmidrule(lr){6-7}
& &
& \textbf{Score} & \textbf{Best R.}
& \textbf{Score} & \textbf{Best R.} \\
\midrule[0.8pt]

Single-Agent Inference
& ReAct $_{\texttt{Pass@1}}$ & Gemini-3-Flash
& 28.6 & --
& 21.8 & -- \\

\midrule[0.3pt]

\multirow{5}{*}{\makecell{Procedure-Based\\Evolution}}
& ADAS  & Gemini-3-Flash
& 38.6 & 7
& \underline{36.0} & 3 \\
& DGM   & Gemini-3-Flash
& \underline{44.3} & 19
& 29.8 & 5 \\
& AFlow & Gemini-3-Flash
& \underline{44.3} & 11
& 31.8 & 14 \\
& SPO   & Gemini-3-Flash
& 42.9 & 19
& 25.0 & 6 \\
& GEPA  & Gemini-3-Flash
& 41.4 & 15
& 22.5 & 13 \\

\midrule[0.3pt]

Ours
& \ourmethod$_{\texttt{Procedure}}$ & Gemini-3-Flash
& \textbf{53.8} & 7
& \textbf{47.0} & 12 \\

\bottomrule[1.2pt]
\end{tabular}

\vspace{0.4em}
\end{table*}

\textbf{Overall performance.}
Tables~\ref{tab:open-ended-results} and~\ref{tab:standard-results} show that \ourmethod consistently improves agentic evolution across both open-ended optimization and fixed benchmarks.
On the open-ended tasks, \ourmethod achieves the best or tied-best result on all three tasks, while also improving the speed or stability with which strong candidates are found.
In particular, on the Kernel optimization task, \ourmethod achieves 1138 cycles within 100 iterations, which is, to our knowledge, the best reported result under the same iteration budget.
This suggests that the harnessed meta-editing loop improves how evolution uses feedback over time, rather than merely increasing the number of candidate attempts.
On standard agentic and reasoning benchmarks, \ourmethod also improves procedure-based evolution over strong fixed-loop baselines.
The gains are consistent across both Terminal-Bench and ARC-AGI-2, yielding a 26\% relative improvement over the strongest baseline on average.
Together, these results support the central claim that mechanism-level intervention can benefit both open-ended optimization and benchmark-driven agentic evolution.

\textbf{Improvement through optimization-time reasoning.}
The gains on Table \ref{tab:standard-results} come with a higher per-round optimization cost.
\ourmethod costs about three times as much as procedure-based baselines on these benchmarks.
This means that \ourmethod improves performance by scaling the reasoning and deliberation used during optimization, and supports our view that increasing the budget of the evolution process is a useful axis for improving agentic evolution.

\textbf{Cost analysis and agentic behavior.}
The open-ended tasks reveal that cost is not solely determined by whether a method is agent-based or procedure-based.
Agent-based evolution can remain cost-competitive when implemented through coding-agent interfaces with prompt caching and persistent contexts.
\ourmethod$_{\texttt{Agent}}$ maintains low per-round cost: 0.34--0.32 on \textit{circle\_packing\_26}, 1.40--1.31 on \textit{autocorrelation\_second}, and 1.27--1.23 on Kernel optimization.
By contrast, procedure-based methods can become expensive in long-horizon optimization by repeatedly constructing large prompts over an expanding search history without comparable caching, as visible in HyperAgents' higher per-round cost on Task 1 and Task 2.

At the same time, direct coding agents show why agent freedom alone is insufficient for reliable evolution.
Even with prompts encouraging long-horizon search, coding agents often stop early once local improvements become difficult, as seen in the early best rounds of Codex on Task 1 and Task 3.
This suggests that an agent's internal stopping decision can conflict with the external evolution budget.
\ourmethod addresses this by placing the coding agent inside an explicit evolution harness, where rounds, candidate records, and evaluation feedback are maintained outside the agent's local decision loop.

\subsection{Evolution Dynamics}

\begin{figure}[htbp]
\centering
\includegraphics[width=\linewidth]{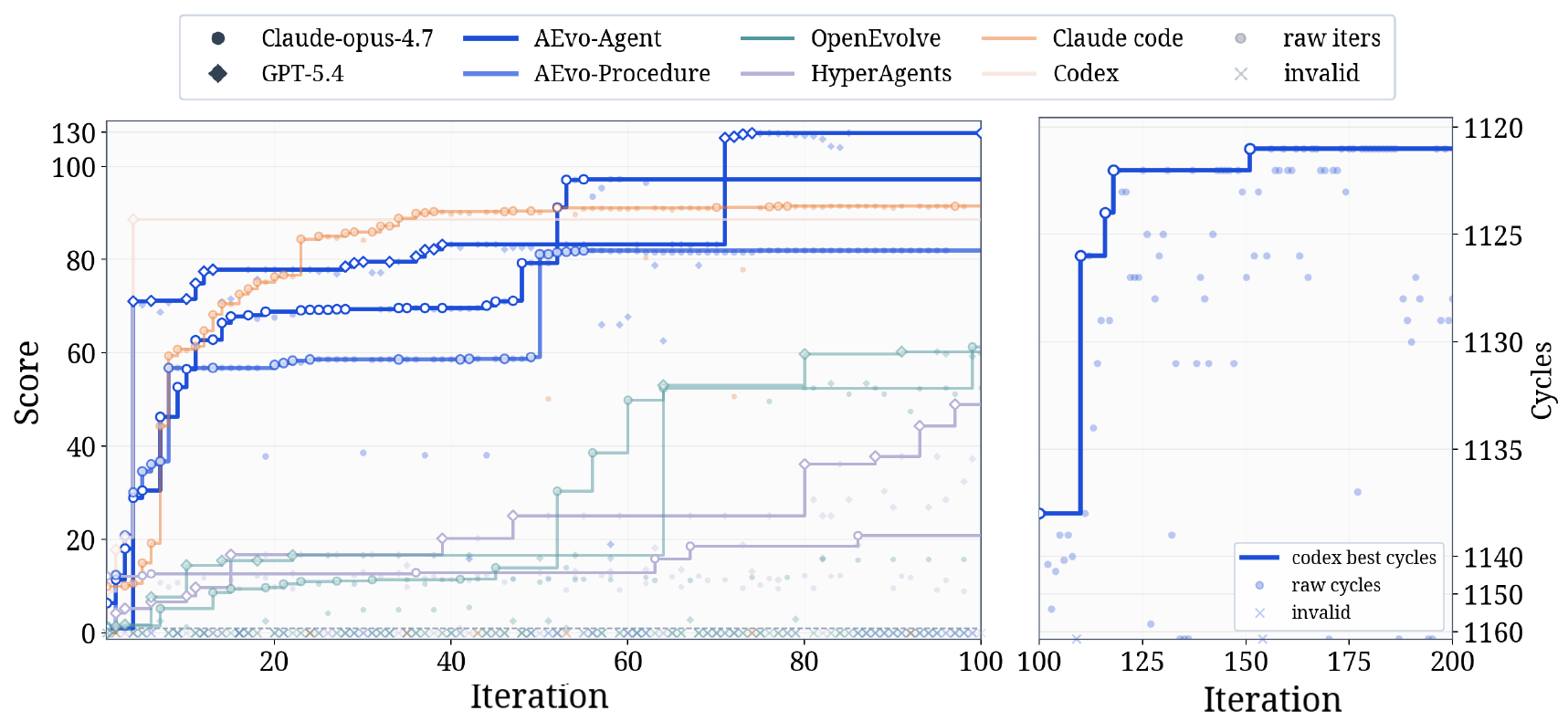}
\caption{
Evolution trajectories on the \textit{Kernel optimization} task.
The left panel compares eight methods over the first 100 iterations, where blue curves denote \ourmethod variants.
The y-axis reports the normalized score induced by cycle reduction, so higher is better; raw iterations and invalid evaluations are shown as scattered markers.
The right panel extends the \ourmethod run from 100 to 200 iterations and reports raw cycles, where lower is better.
}
\label{fig:evolution-dynamics}
\end{figure}

\textbf{Revising evolution after plateaus.}
As shown in Figure~\ref{fig:evolution-dynamics}, procedure-based methods such as OpenEvolve and HyperAgents tend to flatten once their current selection or mutation strategy stops producing useful candidates.
In contrast, \ourmethod can revise the mechanism that drives subsequent evolution.
When progress stalls or repeated failures appear, these signals become process-level feedback: the meta-agent can adjust the procedure, directive, or reusable search context, producing step-wise improvements after plateaus.
This is visible in the late-stage jump that leads to the best 100-round result.

\textbf{Using the evolution budget effectively.}
Direct coding agents can obtain strong early gains through internal simulation, execution, and debugging, but they may stop early once local progress becomes difficult.
\ourmethod avoids tying progress to the agent's local stopping decision by maintaining explicit rounds, candidate records, and evaluation feedback outside the agent context.
This allows the external evolution budget to be used more consistently.

\textbf{Scaling beyond early gains.}
The right panel extends Codex-based \ourmethod from 100 to 200 iterations.
The best result improves from 1138 to 1121 cycles, showing that \ourmethod continues to benefit from additional rounds rather than saturating after an early strong candidate.
Overall, the trajectories suggest that agent flexibility is useful, but reliable long-horizon improvement requires a harness that preserves global evidence and enables mechanism-level correction.

\subsection{Ablation Study}

Table~\ref{tab:ablation} (Appendix) ablates two key components of \ourmethod$_{\texttt{Agent}}$ on the Kernel optimization task. The full system completes the 100-round budget without reward hacking and reaches the best valid result of 1138 cycles. Removing meta-agent skills does not lead to reward hacking, but it substantially weakens long-horizon search: the best run only reaches 1407 cycles, and the runs do not consistently sustain the full budget. Removing the evolution harness is even less reliable. Although one run finds a strong 1167-cycle solution, two of the three runs enter reward-hacking trajectories and fail to produce valid cycle results. These results suggest that the skills mainly support sustained and effective meta-intervention, while the harness provides the protected evaluation boundary and structured evolution context needed to keep agentic search aligned with the true objective.

\subsection{Case Study}

\begin{figure}[t]
    \centering
    \includegraphics[width=1\linewidth]{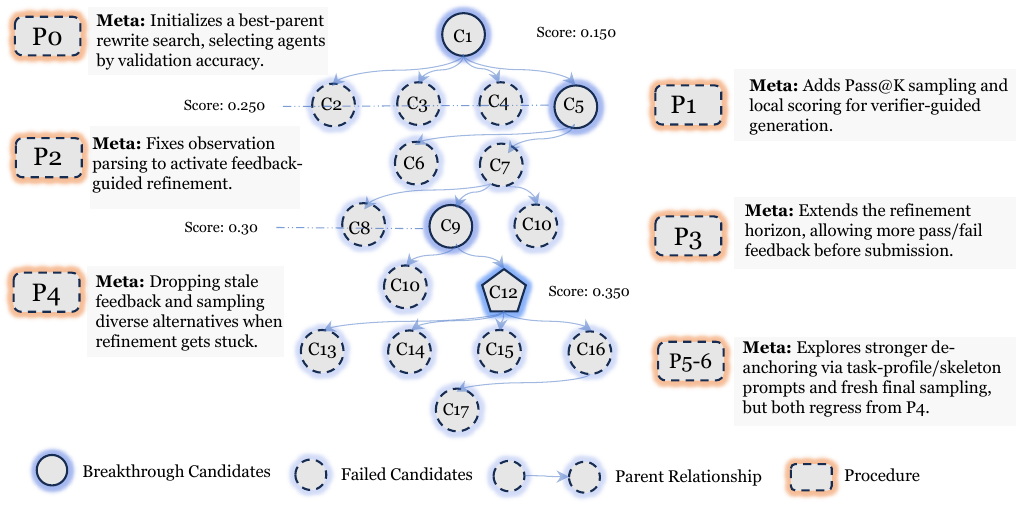}
    \caption{
    Case study of procedure evolution on an ARC-AGI-2 task.
    Each $P$ denotes a procedure produced or revised by the meta-agent, and each $C$ denotes a candidate agent generated by the current procedure.
    Solid nodes are breakthrough candidates, dashed nodes are failed candidates, and arrows indicate parent relationships.
    The meta-agent improves the search process by changing the procedure across stages, while failed candidates provide feedback for later interventions.
    }
    \label{fig:case-study}
\end{figure}

Figure~\ref{fig:case-study} illustrates how \ourmethod performs meta-intervention during procedure-based evolution on an ARC-AGI-2 task.
Starting from $P_0$, the meta-agent initializes a best-parent rewrite procedure that selects candidate agents by validation accuracy.
This first produces an initial breakthrough candidate $C_1$, but subsequent variants expose several failure modes in observation parsing and refinement.
The meta-agent then revises the procedure rather than continuing the same search blindly: $P_1$ adds Pass@K sampling and local scoring for verifier-guided generation, $P_2$ fixes observation parsing to activate feedback-guided refinement, and $P_3$ extends the refinement horizon to use more pass/fail feedback before submission.
When the search becomes stuck, $P_4$ drops stale feedback and samples more diverse alternatives, leading to a stronger candidate.
Later interventions $P_{5}$--$P_{6}$ explore stronger de-anchoring through task-profile and skeleton prompts, but these regress from $P_4$.
This example shows that failed candidates are not merely discarded; they become process-level evidence that helps the meta-agent decide how to revise the future evolution procedure.
Additional evolved procedures, optimized agent harnesses, and task-level analyses are provided in Appendix~\ref{appendix:opt_procedure} and Appendix~\ref{appendix:opt_harness}.

\section{Conclusion}

We presented \method, a harnessed framework for steering agentic evolution by treating the evolution process itself as an interactive environment. Instead of generating one more candidate, \method exposes accumulated candidates, feedback, traces, failures, costs, and search history as process-level evidence, and uses a meta-agent to edit the mechanism that controls future evolution. This formulation provides a unified view of procedure-based and agent-based evolution: the same meta-editing loop can revise explicit search procedures or the operating context of general-purpose evolution agents, while keeping evaluation and candidate recording protected by an external harness. Across agentic, reasoning, and open-ended optimization tasks, \method improves over strong fixed-procedure and agent-based baselines, suggesting that long-horizon evolution benefits not only from stronger candidate generators, but also from mechanism-level intervention over how search proceeds. Future work should study more diverse evolution environments, cheaper meta-intervention strategies, and safer deployment of harnessed agentic evolution for scientific discovery, software engineering, and autonomous code optimization.

\newpage
\bibliography{cited}
\bibliographystyle{plainnat}

\newpage
\appendix
\section{Ablation Study Details}
\label{appendix:ablation}

\begin{table}[h]
\caption{
Ablation study on the Kernel optimization task.
Full reports the main \ourmethod$_{\texttt{Agent}}$ setting, while each ablation reports three independent runs.
Total R. denotes completed evolution rounds, and Best R. denotes the first round reaching the reported best cycles.
}
\label{tab:ablation}
\centering
\small
\setlength{\tabcolsep}{3.6pt}
\renewcommand{\arraystretch}{1.15}

\begin{tabular}{llccccc}
\toprule[1.2pt]
\textbf{Method}
& \textbf{Run}
& \textbf{Reward Hack}
& \textbf{Total R.}
& \textbf{Cycles $\downarrow$}
& \textbf{Best R.}
& \textbf{Invalid / Total} \\
\midrule[0.8pt]

Full
& --
& No
& 100
& \textbf{1138}
& 99
& 16/100 \\

\midrule[0.3pt]

\multirow{3}{*}{\makecell[l]{w/o Meta-Agent\\Skills}}
& 1 & No & 37 & 2379 & 18 & 2/37 \\
& 2 & No & 99 & 1536 & 99 & 28/99 \\
& 3 & No & 65 & 1407 & 53 & 21/65 \\

\midrule[0.3pt]

\multirow{3}{*}{\makecell[l]{w/o Evolution\\Harness}}
& 1 & No  & 100 & 1167 & 81 & 19/100 \\
& 2 & Yes & 100 & N/A  & N/A & 47/100 \\
& 3 & Yes & 57  & N/A  & N/A & 22/57 \\

\bottomrule[1.2pt]
\end{tabular}
\end{table}

\section{Implementation Details}
\label{appendix:impl_details}

When temperature is exposed, we set it to 1; when reasoning-effort control is available, we use the high setting; and we use a maximum context budget of 128k tokens. We run 20 optimization rounds on Terminal-Bench and ARC-AGI-2. On open-ended optimization tasks, we allow up to 100 rounds and early-stop when a run reaches the known target score or fails to improve for 25 consecutive rounds. The initial procedure in the procedure-based setting uses best-valid-candidate selection plus a heuristic LLM optimizer; details are given in Appendix~\ref{appendix:init_procedure}. The initial agent used on the standard agentic and reasoning tasks is a ReAct-style~\citep{yao2022react} agent; details are given in Appendix~\ref{appendix:init_agent}.

\section{Additional Experimental Details}
\label{appendix:details}

\subsection{Task and Metric Details}
\label{appendix:task_metric_details}

Our main experiments cover two standard benchmarks and three open-ended optimization tasks. Terminal-Bench evaluates end-to-end task completion in terminal environments, while ARC-AGI-2 measures abstract reasoning under fixed evaluation rules. The three open-ended tasks use hidden or fixed external evaluators and require the optimizer to improve executable code rather than only produce one-shot answers.

\paragraph{Circle Packing.}
In \textit{circle\_packing\_26}, the goal is to pack 26 circles into a unit square and maximize the sum of their radii. The evaluator returns a validity bit and the achieved packing score. Higher is better.

\paragraph{Autocorrelation-Second.}
In \textit{autocorrelation\_second}, the goal is to construct a non-negative function on $[-\nicefrac{1}{4}, \nicefrac{1}{4}]$ that maximizes
\[
R(f)=\frac{\|f*f\|_2^2}{\|f*f\|_1 \cdot \|f*f\|_\infty}.
\]
The evaluator scores the submitted construction directly by this ratio. Higher is better.

\paragraph{Performance Engineering.}
In the performance-engineering take-home, the goal is to optimize a kernel for a simulated VLIW SIMD machine while preserving correctness on the hidden tests. We report raw cycle count in the main table, so lower is better, although the evaluator also exposes a normalized score derived from the cycle count.

\paragraph{Reporting Protocol.}
For Terminal-Bench and ARC-AGI-2, we report Avg@3 across three independent runs, together with the first optimization round that reaches the best score. For open-ended optimization, we report Best@3 across three runs under a fixed evaluation budget, together with the first round that reaches the best result and the average dollar cost per optimization round. Open-ended runs are capped at 100 rounds and are early-stopped if they reach the known target or plateau for 25 consecutive rounds.

\paragraph{Cost Computation.}
For a run with $R$ optimization rounds, the average dollar cost per round is
\[
\$/R=\frac{1}{R}\sum_{r=1}^{R}\left(
p_{\mathrm{in}} n^{(r)}_{\mathrm{in}}
+
p_{\mathrm{cache}} n^{(r)}_{\mathrm{cache}}
+
p_{\mathrm{out}} n^{(r)}_{\mathrm{out}}
\right),
\]
where $n^{(r)}_{\mathrm{in}}$, $n^{(r)}_{\mathrm{cache}}$, and $n^{(r)}_{\mathrm{out}}$ denote the input, cached-input, and output token counts in round $r$, and $p_{\mathrm{in}}$, $p_{\mathrm{cache}}$, and $p_{\mathrm{out}}$ are the corresponding provider prices.

\subsection{Initialization Details}

\subsubsection{Initial Procedure in Procedure-Based \method}
\label{appendix:init_procedure}

The initial procedure used by procedure-based \ourmethod is intentionally minimal. It selects the current best valid candidate as the parent, applies a single LLM rewrite step, and then invokes the fixed evaluator. The excerpt below shows the corresponding editable surface.

\begin{lstlisting}[style=papercode,language=Python]
class BestByScoreSelection(Selection):
    async def select(self, harness: Harness) -> InfoBundle:
        latest = harness.latest()
        if latest is None:
            seed = harness.new_candidate(parent_round=None)
            harness.write_artifact(
                seed, SOLUTION_FILE, _read(harness.workspace / SEED_FILE)
            )
            seed_metrics = await ReadEvalEvaluation().evaluate(seed)
            harness.record_metrics(seed.round, seed_metrics)
            latest = seed

        best = harness.best(metric="combined_score") or latest
        return InfoBundle(parent=best, notes=f"best round={best.round}")


class LLMRewriteOptimization(Optimization):
    SYSTEM_PROMPT = (
        "You are an expert Python programmer optimizing a circle-packing "
        "solution. Respond with ONLY a single ```python ...``` block."
    )

    async def optimize(self, info, harness, new_candidate):
        parent_code = _read(info.parent.artifacts_dir / SOLUTION_FILE)
        parent_metrics = harness.read_metrics(info.parent.round)
        prompt = (
            f"Current program (score={parent_metrics.get('combined_score', 'n/a')}):\n"
            f"```python\n{parent_code}\n```"
        )
        raw = await llm(prompt, max_tokens=128000)
        harness.write_artifact(
            new_candidate, SOLUTION_FILE, _extract_python_block(str(raw))
        )
\end{lstlisting}

This initialization exposes a limited search surface: it uses score-based selection, a single-parent rewrite step, and no richer failure analysis. This makes later meta-edits easier to interpret because they modify these exposed handles rather than the evaluator.

\subsubsection{Initial Agent in Agentic and Reasoning Tasks}
\label{appendix:init_agent}

For Terminal-Bench and ARC-AGI-2, the initial agent is a minimal ReAct-style agent. The prompt surface and the `step()` loop are both editable, but the seed version contains only basic reasoning, short-horizon memory, and a strict JSON+`bash` action protocol.

\begin{lstlisting}[style=papercode,language=Python]
REACT_PROMPT = """
==== Instruction ====
{instruction}

==== Action Space ====
{action_space}

==== Memory ====
Recent memory:
{memory}

==== Current Observation ====
{obs}

==== Thinking ====
You should think step by step before you output an action.
"""

async def step(self, observation: Observation, history: Any):
    act_prompt = REACT_PROMPT.format(
        instruction=self.current_env_instruction,
        action_space=self.current_action_space,
        obs=observation,
        memory=self._get_memory(),
    )
    resp = await self.llm(act_prompt)
    memory = parse_llm_output(resp, "memory")
    action = self.parse_action(resp)

    if isinstance(action, dict) and action.get("action") == "execute":
        params = action.setdefault("params", {})
        if not params.get("command"):
            bash_cmd = self._extract_bash_command(resp)
            if bash_cmd:
                params["command"] = bash_cmd

    await self.memory.add_memory(obs=agent_obs, action=action, thinking=memory)
    return action, resp, act_prompt
\end{lstlisting}

This weakly structured initialization makes subsequent improvements easier to attribute to changes in prompts, memory management, recovery logic, or context organization rather than to a highly engineered seed agent.

\subsubsection{Meta-Agent Skill Excerpt}
\label{app:meta_agent_skill}

The meta-agent is governed by a compact operational skill that constrains how it reads the workspace, attributes failure, and chooses one causal intervention at a time. A representative excerpt is shown below.

\begin{lstlisting}[style=papercode]
## 1. Core Loop

Read -> Attribute -> Choose Action -> Run Inner-Agent -> Record

## 4. Choose One Action

At each inter-session boundary, choose exactly one action.

### A. goal change
Change only `sessions/_next_goal.md`.

### B. Harness change
Change durable support files such as:
- `skill/evolve_skill.md`
- `shared/validators/`
- `shared/tools/`
- `shared/notes/`

## 8. Hard Rules

Allowed:
- write `sessions/_next_goal.md`
- edit `skill/evolve_skill.md`
- launch inner-agent through `python -m evolver.cli run-inner-agent`

Forbidden:
- edit `candidates/`
- call `oer-eval eval` directly
- bypass the gateway or evaluator
\end{lstlisting}

The meta-agent therefore acts as a controller over future search rather than as an additional task-facing worker. The skill keeps that role boundary explicit.

\subsection{Representative Evolved Artifacts and Outcomes}

\subsubsection{Representative Evolved Procedures}
\label{appendix:opt_procedure}

The excerpt below is taken from a high-performing ARC-AGI-2 procedure. In later rounds, the evolved procedure forwards not only the current best artifact but also per-task slices from alternative references, enabling the optimizer to formulate a single causal hypothesis from those diagnostics.

\begin{lstlisting}[style=papercode,language=Python]
class CrossCandidateSelection(Selection):
    REF_K = 2

    async def select(self, harness: Harness) -> InfoBundle:
        ...
        scored.sort(key=lambda t: (t[1], t[0].round), reverse=True)

        best_cand, best_acc, _ = scored[0]
        refs = []
        ref_notes = []
        for c, acc, _metrics in scored[1:]:
            if len(refs) >= self.REF_K:
                break
            refs.append(c)
            tasks_block = _per_task_results_block(
                _read(c.folder / "eval_output.txt")
            )
            ref_notes.append(
                f"  Round {c.round}: accuracy={acc:.4f}\n"
                f"{_indent(tasks_block, '    ')}"
            )

        best_tasks = _per_task_results_block(
            _read(best_cand.folder / "eval_output.txt")
        )

        notes_lines = [
            f"Selection: best parent round={best_cand.round} accuracy={best_acc:.4f}",
            f"Total candidates so far: {len(all_rounds)}",
            "",
            "[Best per-task results]",
            _indent(best_tasks or "(no per-task block)", "  "),
        ]
        return InfoBundle(
            parent=best_cand,
            references=refs,
            notes="\n".join(notes_lines),
        )
\end{lstlisting}

\begin{lstlisting}[style=papercode,language=Python]
class LLMRewriteAgentOptimization(Optimization):
    async def optimize(self, info, harness, new_candidate):
        parent_code = _read(info.parent.artifacts_dir / ARTIFACT_FILE)
        parent_metrics = harness.read_metrics(info.parent.round)
        parent_eval_log = _read(info.parent.folder / "eval_output.txt")

        parent_acc = float(parent_metrics.get("accuracy", 0.0) or 0.0)
        ref_blocks = []
        for ref in info.references or []:
            ref_code = _read(ref.artifacts_dir / ARTIFACT_FILE)
            ref_metrics = harness.read_metrics(ref.round) or {}
            ref_acc = float(ref_metrics.get("accuracy", 0.0) or 0.0)
            label = "ANTI-EXAMPLE" if ref_acc < parent_acc else "REFERENCE"
            ref_blocks.append(
                f"### {label} round {ref.round} "
                f"(accuracy={ref_acc:.4f})\n"
                f"```python\n{ref_code[:6000]}\n```"
            )
        ref_section = "\n\n".join(ref_blocks) if ref_blocks else "(none yet)"

        base_prompt = (
            f"## Selection notes\n{info.notes}\n\n"
            f"## Current agent.py "
            f"(round {info.parent.round}, accuracy={parent_metrics.get('accuracy', 'n/a')})\n"
            f"```python\n{parent_code}\n```\n\n"
            f"## Recent eval output (tail)\n"
            f"```\n{parent_eval_log[-_EVAL_LOG_TAIL:]}\n```\n\n"
            f"## Alternative references\n{ref_section}\n\n"
            f"## Task\n"
            f"Produce an improved agent.py. Form ONE specific theory "
            f"about the current failures and test the smallest edit "
            f"that directly probes that theory."
        )
\end{lstlisting}

This excerpt illustrates the main leverage of procedure-mode \ourmethod: evolution changes the evidence presented to the optimizer and the way that evidence is structured, not only the candidate artifact being scored.

\subsubsection{Representative Evolved Agent Harness}
\label{appendix:opt_harness}

In the agent-based setting, the evolved object is the harness seen by future inner-agent sessions rather than a single submitted artifact. In the performance-engineering run, this harness contained at least five durable layers: a task skill, a session-specific goal, a persistent family map, support utilities for evaluation accounting, and structured session notes written back into the workspace.

\paragraph{Task skill.}
\begin{lstlisting}[style=papercode]
# Evolve Skill - Performance Engineering (VLIW Kernel)

Optimize a kernel on a simulated VLIW SIMD architecture to minimize
clock cycles. Combined Score = `147734 / cycles` (higher is better;
the goal at the top of your
prompt pins this session's focus - read it before reading anything
else.

## Eval

$ oer-eval eval --program attempts/v1.py
Combined Score: 6.38
Validity:       1.0
Remaining Evals: 95
Session Evals Remaining: 9

## Don't stop early

While `Session Evals Remaining > 0` you do not get to exit. "Current
cycle count looks good" is not a stop condition; an unspent eval is a
planned experiment you failed to run.

If you feel done, that is the signal to read
`candidates/candidate_<best>/program.py` (or your own best vN.py),
name the specific structural reason it's stuck, and submit a candidate
from a different family.

## CRITICAL - eval DB and the session-2 finding

1. Stage the local DB once at session start:
   `cp shared/notes/oer_eval_local_template.db ./.oer_eval.local.db`
2. Run all evals with an explicit local DB:
   `oer-eval eval --program attempts/vN.py --db-path ./.oer_eval.local.db`
3. The meta agent will replay the rows from
   `./.oer_eval.local.db` into the workspace `.oer_eval.db` after
   your session ends.
4. Do NOT spend evals re-confirming the readonly issue.
5. NEVER copy or mutate `../../../.oer_eval.db` directly from
   inside the sandbox.

## SESSION_NOTES.md (required on exit)

Write `SESSION_NOTES.md` at your cwd root before finishing.
\end{lstlisting}

\paragraph{Session goal.}
\begin{lstlisting}[style=papercode]
# goal for next inner-agent session (session 7)

## Status
- Current best: 1774 cycles, score 83.28.
- 61 evals remaining in the global quota; this session has MAX_EVALS = 15.
- load is now the dominant bottleneck again.
- Session 6 only used 6 of 15 evals before exiting.

## Setup commands
cp shared/notes/oer_eval_local_template.db ./.oer_eval.local.db
mkdir -p attempts
cp shared/notes/best_v59_session6.py attempts/v59_parent.py
oer-eval eval --program attempts/v59_parent.py --db-path ./.oer_eval.local.db

## Hypotheses (in test order - DO NOT skip later ones if you finish early)

Test 1: family D'' - depth-3 cache.
Test 2: family C - software-pipelined inner loop.
Test 3: hash-chain cross-stage algebra.
Test 4: scheduler tie-breaks aware of engine occupancy.
Test 5: aggressive scratch reclaim + bundle merging.

## Required behavior
- MAX_EVALS = 15. Submit 15 attempts.
- Use `--db-path ./.oer_eval.local.db` on every eval.
- Save attempts as `attempts/vN.py`.
- Use offline `Engine` simulator for sanity before eval.

## Required output (`SESSION_NOTES.md`)
- Best Combined Score and corresponding `attempts/vN.py`.
- v59 starting profile and the v_best ending profile.
- For each attempt: score, validity, structural change, which engine shifted.
- Final local-DB `Remaining Evals`.
- One specific, falsifiable hypothesis for session 8.
\end{lstlisting}

\paragraph{Persistent family map.}
\begin{lstlisting}[style=papercode]
# Family Map - VLIW Kernel

## Architecture cheatsheet
- VLEN=8, batch_size=256 -> 32 SIMD groups
- rounds=16
- SLOT_LIMITS: alu 12 / valu 6 / load 2 / store 2 / flow 1
- 6 hash stages, each pure valu (`+`, `^`, `<<`, `>>`)
- Per element: parity -> next_idx = 2*idx + (1 if even else 2)

## Sessions
### session 4 - family B+C: fused hash + ping-pong + cached depth-1 reuse
- best 77.88 / 1897 cycles (`v31`)
- v27: collapse hash stages 0/2/4 into `multiply_add`
- v31: cached depth-1 node `vselect` across rounds

### session 5 - family D: idx-update structural reductions
- best 79.60 / 1856 cycles (`v47`)
- stage-5 short-circuit hypothesis was FALSIFIED
- v43: depth-1 base preselection
- v47: generic non-root precompute of `2*idx + 1`

### session 9 - breakthrough: 1140 cycles via explicit-family port
- best 129.59 / 1140 cycles (`v91`)
- v88 explicit-family port (-597)
- v89 depth-1 madd (-10)
- v90 depth-3 2-madd (-19)
- v91 scalar alu parity + per-lane xor (-8)

## Do not repeat
- Scheduler tie-break / priority tweaks alone.
- Stage-5 short-circuit `(a^const)^(a>>16)`.
- Stage-4 -> stage-5 fusion via `multiply_add`.

## Workflow gotchas
- Use `--db-path ./.oer_eval.local.db` after copying
  `shared/notes/oer_eval_local_template.db`.
- Meta agent replays after session ends.
\end{lstlisting}

\paragraph{Replay utility.}
\begin{lstlisting}[style=papercode,language=Python]
#!/usr/bin/env python3
"""Replay rows from a session's local oer-eval DB into the workspace DB.

Used by the meta agent ONLY (after a session ends) to credit evals that the
codex sandbox forced into a session-local DB. Inner agents must not run this.
"""
session_id = int(sys.argv[1])
ws = Path(__file__).resolve().parent.parent.parent
session_dir = ws / "sessions" / f"session_{session_id}"
local_db = session_dir / "agent_workspace" / ".oer_eval.local.db"
ws_db = ws / ".oer_eval.db"

src_rows = src.execute(
    "SELECT agent_name, problem_name, program_path, program_content, "
    "validity, eval_time, combined_score, error, raw_result "
    "FROM evaluations ORDER BY id"
).fetchall()

for r in src_rows:
    dst.execute(
        "INSERT INTO evaluations "
        "(agent_name, problem_name, program_path, program_content, "
        " validity, eval_time, combined_score, error, raw_result) "
        "VALUES (?, ?, ?, ?, ?, ?, ?, ?, ?)",
        (agent_name, r[1], r[2], r[3], r[4], r[5], r[6], r[7], r[8]),
    )
\end{lstlisting}

\paragraph{Session memory written back by the inner agent.}
\begin{lstlisting}[style=papercode]
# Session Notes - session 9

## Best result
- Best Combined Score: 129.59
- Best cycles: 1140
- Best file: `attempts/v91.py`

## Attempt log
- `attempts/v88.py` - explicit-family port
- `attempts/v89.py` - `depth1` node select `vselect -> madd`
- `attempts/v90.py` - reduced-flow `depth3` selector
- `attempts/v91.py` - scalar root parity + per-lane alu xor
- `attempts/v97.py` - full depth-2 two-`madd` rewrite
- `attempts/v98.py` - scalarize generic-round post-hash parity extraction

## What worked
- `v88` cut 597 cycles from `v59_parent`.
- `depth1` binary-select to `madd` paid.
- `depth3` reduced-flow selector paid more.

## What did not work
- naive generic-round selector rewrite was wrong for this schedule shape.
- fully flow-free depth-2 rewrite also hurt.
- scalarizing generic-round parity was catastrophic.

## Hypothesis for session 10
- a partial depth-2 rewrite that replaces only one of the two
  `flow.vselect`s with `madd` should beat 1140.
\end{lstlisting}

Together, these excerpts show that the evolved agent harness is not a single prompt edit. It is a layered control structure consisting of persistent instructions, hypothesis-carrying goals, accumulated family-level memory, support code for evaluator interaction, and structured records that are promoted into future sessions.

\subsubsection{Representative Optimization Outcomes}
\label{appendix:opt_results}

\paragraph{ARC-AGI-2 best artifact.}
The best ARC-AGI-2 artifact in our run reaches accuracy $0.35$ ($7/20$). Structurally, the final agent has three persistent components: a prompt surface that distinguishes normal feedback refinement from ``fresh exploration,'' a local Pass@K scorer over cached training pairs, and a validate-versus-submit controller that uses the best local score and plateau state.

\begin{lstlisting}[style=papercode,language=Python]
FEEDBACK_SUFFIX = """
==== Previous Attempt Feedback ====
A previous attempt produced this code:
```python
{prev_code}
```
It passed {passed}/{total} training examples.

FAILING examples (must fix these):
{fail_details}
"""

FRESH_EXPLORE_SUFFIX = """
==== Fresh Exploration Required ====
Previous attempts have been stuck at a partial solution.
IGNORE any previous approach and propose a completely
different interpretation of the transformation rule.
"""

DIVERSITY_HINTS = [
    "",
    "Consider symmetry, rotation, reflection, or tiling patterns.",
    "Consider connected components, object counting, or shape detection.",
    "Consider color mapping, replacement, or color-dependent rules.",
]

class ArcAGIAgent(BaseAgent):
    """Pass@K with fresh-exploration escape when stuck at local optimum."""

    num_samples: int = Field(default=3)
    cached_train_pairs: List[Dict[str, Any]] = Field(default_factory=list)
    best_score_so_far: int = Field(default=-1)
    best_response_so_far: str = Field(default="")
    best_code_so_far: str = Field(default="")
    best_fails: List[Tuple[int, Any]] = Field(default_factory=list)
    hint_rotation_offset: int = Field(default=0)
    stuck_counter: int = Field(default=0)
\end{lstlisting}

\begin{lstlisting}[style=papercode,language=Python]
class ArcAGIAgent(BaseAgent):
    async def step(self, observation: Observation, history: Any):
        ...
        base_prompt = ARCAGI_PROMPT.format(
            instruction=instruction_text,
            action_space=self.current_action_space,
            memory=self.memory.as_text() if self.memory else "None",
            obs=observation,
        )

        train_pairs = _extract_train_pairs(observation)
        if train_pairs:
            self.cached_train_pairs = train_pairs
        elif self.cached_train_pairs:
            train_pairs = self.cached_train_pairs
        n_train = len(train_pairs)

        use_fresh_explore = (
            self.stuck_counter >= 2
            and n_train > 0
            and 0 <= self.best_score_so_far < n_train
        )

        if not use_fresh_explore and (
            n_train > 0
            and self.best_score_so_far >= 0
            and self.best_score_so_far < n_train
            and self.best_code_so_far
            and self.best_fails
        ):
            fail_details = _format_fail_details(self.best_fails, train_pairs)
            feedback_suffix = FEEDBACK_SUFFIX.format(
                prev_code=self.best_code_so_far[:3000],
                passed=self.best_score_so_far,
                total=n_train,
                fail_details=fail_details,
            )
        elif use_fresh_explore:
            feedback_suffix = FRESH_EXPLORE_SUFFIX

        prompts = []
        for i in range(self.num_samples):
            hint_idx = (i + self.hint_rotation_offset) % len(DIVERSITY_HINTS)
            prompts.append(base_prompt + feedback_suffix + DIVERSITY_HINTS[hint_idx])
        responses = await asyncio.gather(*[self._sample_once(p) for p in prompts])

        for r in valid_responses:
            code = _extract_python_block(r)
            score, fails = _score_candidate(code, train_pairs)
            if score > round_best_score:
                round_best_score = score
                round_best_response = r
                round_best_code = code
                round_best_fails = fails

        if round_best_score > self.best_score_so_far:
            self.best_score_so_far = round_best_score
            self.best_response_so_far = round_best_response
            self.best_code_so_far = round_best_code
            self.best_fails = round_best_fails
            self.stuck_counter = 0
        elif 0 <= self.best_score_so_far < n_train:
            self.stuck_counter += 1

        action = self.parse_action(self.best_response_so_far or round_best_response)
        if self.best_score_so_far >= n_train:
            action["action"] = "submit"
        elif step_num >= self.max_refine_steps:
            action["action"] = "submit"
        else:
            action["action"] = "validate"
\end{lstlisting}

\paragraph{Analysis.}
The improvement over the seed agent comes from a tighter coupling between search control and task-local verification. First, caching the training pairs removes a brittle dependency on the observation format at later steps, so local verification remains available throughout the interaction. Second, Pass@K sampling with prompt diversification converts a single-sample ReAct loop into a small search procedure over candidate solvers, with selection driven by observed agreement on the training pairs rather than by the raw LLM output alone. Third, the separation between \texttt{FEEDBACK\_SUFFIX} and \texttt{FRESH\_EXPLORE\_SUFFIX} makes the agent alternate explicitly between exploitation and hypothesis reset: partial but improving candidates are refined through concrete failure-conditioned feedback, whereas persistent plateaus trigger a prompt regime that suppresses anchoring to the current local optimum. Finally, the validate-versus-submit controller ties action choice to the best verified score rather than to the latest response, which reduces premature submission of partially correct programs.

\paragraph{Performance-engineering best artifact.}
The best performance-engineering artifact is a two-file submission. The top-level program is only a wrapper; the schedule-level optimization resides in a benchmark-specialized base that exposes a small set of round-family control points. The final validated artifact reaches $1138$ cycles by combining evaluator-compatible packaging, explicit specialization of the benchmark rounds, and a non-uniform assignment of selector logic across engines.

\paragraph{Submitted wrapper.}
\begin{lstlisting}[style=papercode,language=Python]
from importlib.util import module_from_spec, spec_from_file_location
from pathlib import Path

_SPEC = spec_from_file_location("variant_base", Path(__file__).with_name("variant_base.py"))
_MOD = module_from_spec(_SPEC)
assert _SPEC.loader is not None
_SPEC.loader.exec_module(_MOD)
VariantKernelBuilderBase = _MOD.VariantKernelBuilderBase

class KernelBuilder(VariantKernelBuilderBase):
    DEPTH1_NODE_STYLE_CALLS = ("madd", "vselect")
\end{lstlisting}

\paragraph{Parameterized benchmark family.}
\begin{lstlisting}[style=papercode,language=Python]
BENCH_FOREST_HEIGHT = 10
BENCH_N_NODES = 2 ** (BENCH_FOREST_HEIGHT + 1) - 1
BENCH_BATCH_SIZE = 256
BENCH_ROUNDS = 16
BENCH_FOREST_BASE = 7
BENCH_INP_INDICES_P = BENCH_FOREST_BASE + BENCH_N_NODES
BENCH_INP_VALUES_P = BENCH_INP_INDICES_P + BENCH_BATCH_SIZE
BENCH_TILES = BENCH_BATCH_SIZE // VLEN

class VariantKernelBuilderBase:
    ROOT_SCALAR_AND_CALLS = (True, False)
    DEPTH1_NODE_STYLE_CALLS = ("madd", "madd")
    DEPTH1_POSTHASH_SCALAR_AND_CALLS = (False, False)
    DEPTH2_STYLE_CALLS = ("base", "base")
    DEPTH2_POSTHASH_SCALAR_AND_CALLS = (False, False)
    DEPTH3_STYLE_CALLS = ("madd", "madd")
    DEPTH3_POSTHASH_SCALAR_AND_CALLS = (False, False)
\end{lstlisting}

\paragraph{Core benchmark schedule.}
\begin{lstlisting}[style=papercode,language=Python]
root_scalar_and_calls = self.ROOT_SCALAR_AND_CALLS
depth1_node_style_calls = self.DEPTH1_NODE_STYLE_CALLS
depth1_posthash_scalar_and_calls = self.DEPTH1_POSTHASH_SCALAR_AND_CALLS
depth2_style_calls = self.DEPTH2_STYLE_CALLS
depth2_posthash_scalar_and_calls = self.DEPTH2_POSTHASH_SCALAR_AND_CALLS
depth3_style_calls = self.DEPTH3_STYLE_CALLS
depth3_posthash_scalar_and_calls = self.DEPTH3_POSTHASH_SCALAR_AND_CALLS

def emit_hash(tile_ids):
    for tile in tile_ids:
        self.emit_madd(vals[tile], vals[tile], mul4097_v, addc1_v)
    for tile in tile_ids:
        self.emit_valu(">>", tmp1s[tile], vals[tile], shift19_v)
    for tile in tile_ids:
        for lane in range(VLEN):
            self.emit_alu("^", vals[tile] + lane, vals[tile] + lane, c2_s)
    for tile in tile_ids:
        self.emit_valu("^", vals[tile], vals[tile], tmp1s[tile])
    for tile in tile_ids:
        self.emit_madd(vals[tile], vals[tile], mul33_v, addc3_v)
    for tile in tile_ids:
        self.emit_valu("<<", tmp1s[tile], vals[tile], shift9_v)
    for tile in tile_ids:
        for lane in range(VLEN):
            self.emit_alu("+", vals[tile] + lane, vals[tile] + lane, c4_s)
    for tile in tile_ids:
        self.emit_valu("^", vals[tile], vals[tile], tmp1s[tile])
    for tile in tile_ids:
        self.emit_madd(vals[tile], vals[tile], mul9_v, addc5_v)
    for tile in tile_ids:
        self.emit_valu(">>", tmp1s[tile], vals[tile], shift16_v)
    for tile in tile_ids:
        for lane in range(VLEN):
            self.emit_alu("^", vals[tile] + lane, vals[tile] + lane, c6_s)
    for tile in tile_ids:
        self.emit_valu("^", vals[tile], vals[tile], tmp1s[tile])

def emit_parity_and(dest_vec: int, src_vec: int, use_scalar_and: bool):
    if use_scalar_and:
        for lane in range(VLEN):
            self.emit_alu("&", dest_vec + lane, src_vec + lane, one_v + lane)
    else:
        self.emit_valu("&", dest_vec, src_vec, one_v)

def round_root(use_scalar_and: bool):
    for tile in tiles:
        self.emit_valu("^", vals[tile], vals[tile], root_v)
    emit_hash(tiles)
    for tile in tiles:
        emit_parity_and(idxs[tile], vals[tile], use_scalar_and)

def round_depth1(node_style: str, use_scalar_and: bool):
    for tile in tiles:
        if node_style == "madd":
            self.emit_madd(tmp0s[tile], idxs[tile], node12_diff_v, node1_v)
        elif node_style == "vselect":
            self.emit_vselect(tmp0s[tile], idxs[tile], node2_v, node1_v)
        else:
            raise ValueError(f"unknown depth1 node style: {node_style}")
    for tile in tiles:
        self.emit_valu("^", vals[tile], vals[tile], tmp0s[tile])
    emit_hash(tiles)
    for tile in tiles:
        emit_parity_and(tmp0s[tile], vals[tile], use_scalar_and)
        self.emit_madd(idxs[tile], idxs[tile], two_v, tmp0s[tile])

def round_depth2(style: str, use_scalar_and: bool):
    for tile in tiles:
        self.emit_valu("&", d2_bit0, idxs[tile], one_v)
        self.emit_valu(">>", d2_bit1, idxs[tile], one_v)
        if style == "base":
            self.emit_vselect(tmp0s[tile], d2_bit0, node4_v, node3_v)
            self.emit_vselect(d2_mix, d2_bit0, node41_diff_v, node30_diff_v)
        elif style == "madd_first":
            self.emit_madd(tmp0s[tile], d2_bit0, node43_diff_v, node3_v)
            self.emit_vselect(d2_mix, d2_bit0, node41_diff_v, node30_diff_v)
        elif style == "madd_second":
            self.emit_vselect(tmp0s[tile], d2_bit0, node4_v, node3_v)
            self.emit_madd(d2_mix, d2_bit0, d2_mix_delta_v, node30_diff_v)
        else:
            raise ValueError(f"unknown depth2 style: {style}")
        self.emit_madd(tmp0s[tile], d2_bit1, d2_mix, tmp0s[tile])
        self.emit_valu("^", vals[tile], vals[tile], tmp0s[tile])
    emit_hash(tiles)
    for tile in tiles:
        emit_parity_and(tmp0s[tile], vals[tile], use_scalar_and)
        self.emit_madd(idxs[tile], idxs[tile], two_v, tmp0s[tile])
\end{lstlisting}

\begin{lstlisting}[style=papercode,language=Python]
def round_depth3(style: str, use_scalar_and: bool):
    for tile in tiles:
        self.emit_valu("&", d3_bit0, idxs[tile], one_v)
        self.emit_valu(">>", d3_bit1, idxs[tile], one_v)
        self.emit_valu("&", d3_bit2, d3_bit1, one_v)
        self.emit_valu(">>", d3_bit1, idxs[tile], two_v)
        if style == "madd":
            self.emit_vselect(d3_pair0, d3_bit0, depth3_nodes[1], depth3_nodes[0])
            self.emit_vselect(d3_pair1, d3_bit0, depth3_diff_lo1_v, depth3_diff_lo0_v)
            self.emit_madd(d3_pair0, d3_bit2, d3_pair1, d3_pair0)
            self.emit_vselect(d3_pair1, d3_bit0, depth3_nodes[5], depth3_nodes[4])
            self.emit_vselect(tmp0s[tile], d3_bit0, depth3_diff_hi1_v, depth3_diff_hi0_v)
            self.emit_madd(d3_pair1, d3_bit2, tmp0s[tile], d3_pair1)
            self.emit_vselect(tmp0s[tile], d3_bit1, d3_pair1, d3_pair0)
        elif style == "vselect":
            self.emit_vselect(d3_pair0, d3_bit0, depth3_nodes[1], depth3_nodes[0])
            self.emit_vselect(d3_pair1, d3_bit0, depth3_nodes[3], depth3_nodes[2])
            self.emit_vselect(d3_pair2, d3_bit0, depth3_nodes[5], depth3_nodes[4])
            self.emit_vselect(d3_pair3, d3_bit0, depth3_nodes[7], depth3_nodes[6])
            self.emit_vselect(d3_pair0, d3_bit2, d3_pair1, d3_pair0)
            self.emit_vselect(d3_pair2, d3_bit2, d3_pair3, d3_pair2)
            self.emit_vselect(tmp0s[tile], d3_bit1, d3_pair2, d3_pair0)
        else:
            raise ValueError(f"unknown depth3 style: {style}")
        for lane in range(VLEN):
            self.emit_alu("^", vals[tile] + lane, vals[tile] + lane, tmp0s[tile] + lane)
    emit_hash(tiles)
    for tile in tiles:
        emit_parity_and(tmp0s[tile], vals[tile], use_scalar_and)
        self.emit_madd(idxs[tile], idxs[tile], two_v, tmp0s[tile])
        self.emit_valu("+", idxs[tile], idxs[tile], depth4_base_v)

def round_gather(update_idx: bool):
    for tile in tiles:
        for lane in range(VLEN):
            self.emit_load_offset(tmp0s[tile], idxs[tile], lane)
    for tile in tiles:
        self.emit_valu("^", vals[tile], vals[tile], tmp0s[tile])
    emit_hash(tiles)
    if update_idx:
        for tile in tiles:
            self.emit_valu("&", tmp0s[tile], vals[tile], one_v)
            self.emit_vselect(tmp1s[tile], tmp0s[tile], add_odd_v, add_even_v)
            self.emit_madd(idxs[tile], idxs[tile], two_v, tmp1s[tile])

round_root(use_scalar_and=root_scalar_and_calls[0])
round_depth1(depth1_node_style_calls[0], depth1_posthash_scalar_and_calls[0])
round_depth2(depth2_style_calls[0], depth2_posthash_scalar_and_calls[0])
round_depth3(depth3_style_calls[0], depth3_posthash_scalar_and_calls[0])
round_gather(update_idx=True)
round_gather(update_idx=True)
round_gather(update_idx=True)
round_gather(update_idx=True)
round_gather(update_idx=True)
round_gather(update_idx=True)
round_gather(update_idx=False)
round_root(use_scalar_and=root_scalar_and_calls[1])
round_depth1(depth1_node_style_calls[1], depth1_posthash_scalar_and_calls[1])
round_depth2(depth2_style_calls[1], depth2_posthash_scalar_and_calls[1])
round_depth3(depth3_style_calls[1], depth3_posthash_scalar_and_calls[1])
round_gather(update_idx=True)
\end{lstlisting}

\begin{lstlisting}[style=papercode]
Evaluation:      valid=True, cycles=1138, speedup=129.82x
Status:          success
Problem:         perf_engineering
Combined Score:  129.8189806678383
Validity:        1.0
Eval Time:       31.66170883178711s
\end{lstlisting}

\paragraph{Analysis.}
The low cycle count comes from three coupled changes. First, the file-local \texttt{importlib} wrapper is not cosmetic: without it, the evaluator rejects the whole family because sibling modules are not imported through the workspace package path. The wrapper therefore preserves a modular implementation while keeping the submission evaluator-compatible. Second, the benchmark-specialized base restructures the kernel around the exact round pattern of the benchmark rather than a generic loop. Its control flow is emitted as four specialized top-of-tree rounds, followed by seven gather rounds, followed by the same four specialized rounds and a final gather; this removes generic control overhead where the traversal repeatedly revisits the upper levels of the tree and makes the period-$11$ reuse pattern directly schedulable. Within each specialized round, \texttt{emit\_hash} is emitted phase-major across tiles, which exposes many independent chains to the scheduler and improves overlap between \texttt{alu}, \texttt{valu}, and memory operations. Third, the remaining gains come from engine balancing rather than from further structural refactoring. On this VLIW benchmark, \texttt{flow} has only one slot per cycle, so replacing a binary selector by \texttt{multiply\_add} is profitable only when the induced \texttt{valu} pressure stays below the new bottleneck. The progression recorded in the run data is consistent with this view: the explicit-family port yields the dominant improvement ($-597$ cycles), additional \texttt{madd}-based selectors at depth 1 and depth 3 save another $29$ cycles, an \texttt{alu}-based parity/XOR rebalance saves $8$, and the final artifact gains the last $2$ cycles by reverting only the second specialized depth-1 selector back to \texttt{vselect}. The best artifact is therefore not simply a shorter program; it is a schedule in which specialization, phase ordering, and per-round engine placement are co-tuned to the simulator's slot limits.

Taken together, these artifacts illustrate two modes of durable improvement in \ourmethod: benchmark tasks improve through changes in how the optimizer reasons over failures, whereas open-ended optimization improves through the preservation and recombination of low-level implementation knowledge across many sessions.

\end{document}